%% file: main.tex
\documentclass{article}

\usepackage{PRIMEarxiv}

\usepackage[utf8]{inputenc} 
\usepackage[T1]{fontenc}    
\usepackage{hyperref}       
\usepackage{url}            
\usepackage{booktabs}       
\usepackage{amsfonts}       
\usepackage{nicefrac}       
\usepackage{microtype}      
\usepackage{lipsum}
\usepackage{fancyhdr}       
\usepackage{graphicx}       
\graphicspath{{media/}}     

\input{math_commands.tex}

\usepackage{dsfont}
\usepackage{longtable}
\def\*#1{\mathbf{#1}}
\newcommand{\+}[1]{\boldsymbol{\mathbf{#1}}}

\usepackage{newfloat}
\usepackage{listings}

\usepackage{caption} 
\usepackage{subcaption}
\usepackage{algorithm}
\usepackage{algorithmic}
\usepackage{multirow}
\usepackage{amsmath}
\usepackage{xcolor}
\usepackage{colortbl}
\usepackage{enumitem}
\usepackage{arydshln}
\usepackage{natbib}

\title{The future is different: Large pre-trained language models fail in prediction tasks}

\author{
  Kostadin Cvejoski \\
  Lamarr Institute and Fraunhofer IAIS \\
  Sankt Augustin, Germany\\
  \texttt{kostadin.cvejoski@iais.fraunhofer.de} \\
   \and
  Rams\'es J. S\'anchez \\
  Lamarr Institute and BIT Uni-Bonn \\
  Bonn, Germany\\
  \texttt{sanchez@bit.uni-bonn.de} \\
  \and
  C\'esar Ojeda \\
  Berlin Center for ML and TU Berlin \\
  Berlin, Germany \\
  \texttt{ojedamarin@tu-berlin.de} \\
}

\begin{document}
\maketitle

\begin{abstract}
Large pre-trained language models (LPLM) have shown spectacular success when fine-tuned on downstream supervised tasks. 
Yet, it is known that their performance can drastically drop when there is a \textit{distribution shift} between the data used during training and that used at inference time. 
In this paper we focus on data distributions that naturally change over time and introduce four new \textsc{Reddit} datasets, namely  the \textsc{Wallstreetbets}, \textsc{AskScience}, \textsc{The Donald}, and \textsc{Politics} sub-reddits.
First, we empirically demonstrate that LPLM can display average performance drops of about 88$\%$ (in the best case!) when predicting the popularity of future posts from sub-reddits whose \textit{topic distribution changes with time}.
We then introduce a simple methodology that leverages neural variational dynamic topic models and attention mechanisms to infer temporal language model representations for regression tasks.
Our models display performance drops of only about 40$\%$ in the worst cases (2$\%$ in the best ones) when predicting the popularity of future posts, while using only about 7$\%$ of the total number of parameters of LPLM and providing interpretable representations that offer insight into real-world events, like the GameStop short squeeze of 2021.
\end{abstract}

\keywords{dynamic topic models \and large pre-trained models \and neural variational inference \and large language models \and opinion dynamics}

\section{Introduction}
\input{introduction}
\label{sec:introduction}

\section{Related Work}
\label{sec:related-work}
\input{related_work}

\section{Model}
\label{sec:model}
\input{model}

\section{Dataset and Experimental Setup}
\input{dataset_experimental_setup}

\label{sec:dataset}

\section{Results and Discussion}
\input{results_and_discussion}
\label{sec:results_and_discussion}

\section*{Acknowledgments}
This research has been funded by the Federal Ministry of Education and Research of Germany and the state of North-Rhine Westphalia as part of the Lamarr-Institute for Machine Learning and Artificial Intelligence, LAMARR22B.

\bibliographystyle{abbrvnat}
\bibliography{bibliography}

\clearpage 
\appendix
\input{appendix}

\end{document}

%% file: math_commands.tex
\usepackage{amsmath,amsfonts,bm}

\def\eqref#1{equation~(\ref{#1})}
\def\Eqref#1{Equation~(\ref{#1})}

\def\1{\bm{1}}

\DeclareMathAlphabet{\mathsfit}{\encodingdefault}{\sfdefault}{m}{sl}
\SetMathAlphabet{\mathsfit}{bold}{\encodingdefault}{\sfdefault}{bx}{n}



%% file: introduction.tex
%
%
%
\begin{figure}[t]
\centering
\includegraphics[keepaspectratio,width=0.9\textwidth]{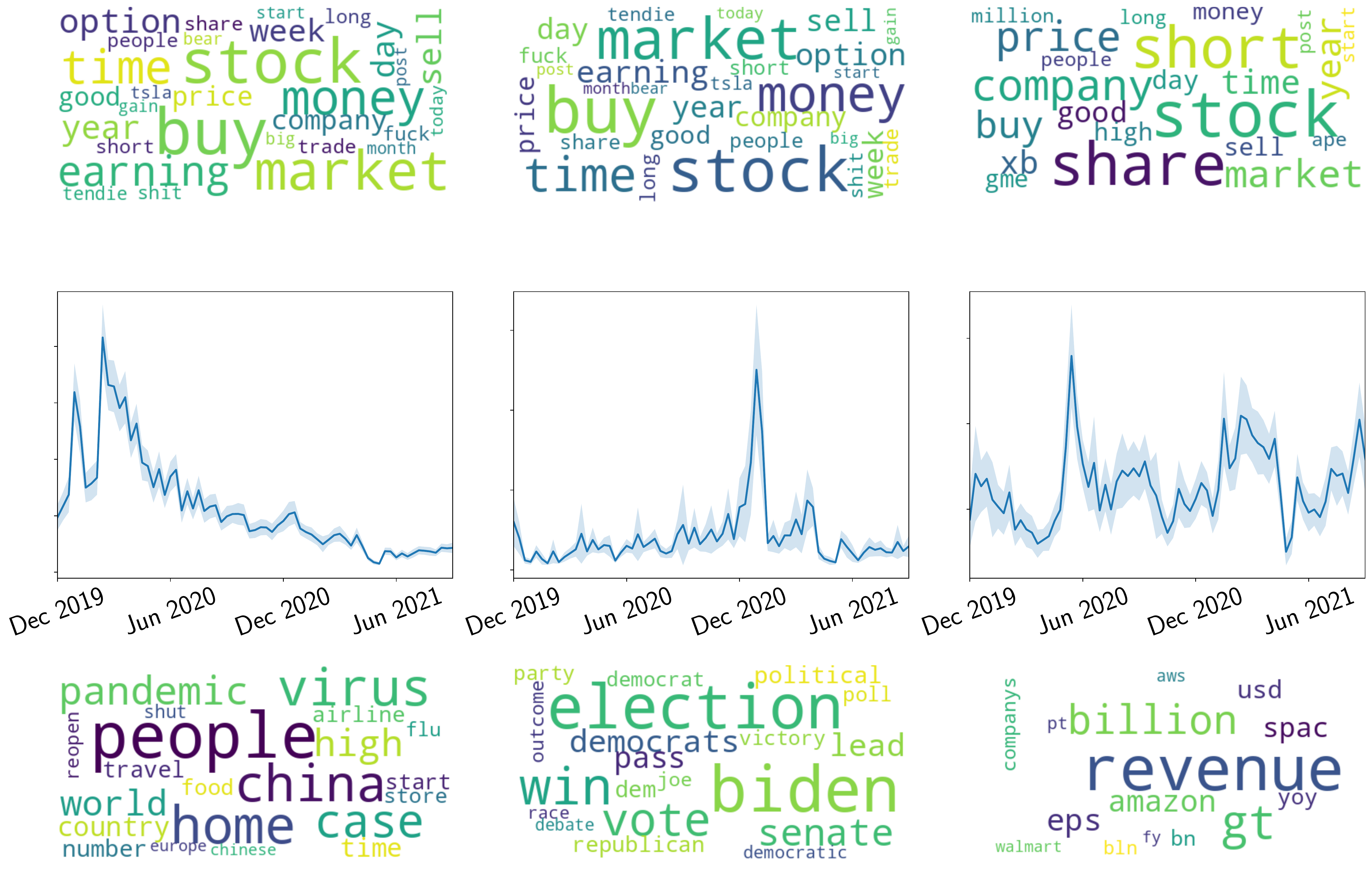} 
\caption{Dynamic features of \textsc{Wallstreetbets}.
Top row: 30 most frequent words for all documents collected within the first, middle and last 8-week windows of the \textsc{Wallstreetbets} datasets.
Middle row: time evolution of topic proportions from three randomly chosen topics, inferred by a neural DTM.
Bottom row: top 30 words associated with each of the topics in the middle row.
The figure illustrates that topic representations, as opposite to simple Bag-of-Word representations, capture (some of) the dynamic components of the dataset.}      
\label{fig:dynamics_of_wallstreetbets}
\end{figure}
The modern natural language processing (NLP) paradigm leverages massive datasets, data-scalable (deep) attention mechanisms and minimal inductive biases \cite{vaswani2017attention} in the form of large pre-trained language models (LPLM) which are subsequently fine-tuned on new learning tasks and datasets \cite{radford2018improving, radford2019language}.
This approach has proven a tremendous success in supervised learning, where applications include question answering, sentiment analysis, named entity recognition and textual entailment, just to name a few (see e.g. \cite{peters-etal-2018-deep, devlin2018bert, liu2019roberta, lan2019albert} and \cite{qiu2020pre} for a review).
%
%
%
Yet, many recent works have also reported the strong sensitivity of LPLM to \textit{distribution shifts} between the data used during training and that
used at inference time \cite{ma-etal-2019-domain, hendrycks-etal-2020-pretrained, zhou-etal-2021-contrastive, chawla2021quantifying}.
In other words, fine-tuned LPLM are known to suffer significantly at zero-shot when applied to different data domains.

In this work we focus on a particular type of natural distribution shift which arises in documents collected over long periods of time. 
Indeed, document collections such as magazines, academic journals, news articles and social media content not only feature trends and themes that change with time, but also employ their language differently as time evolves \cite{10.1145/2488388.2488416}.
LPLM fine-tuned on documents collected up to some given time (i.e. on a given observation time window) might therefore perform poorly when evaluated in future documents, if the latter \textit{differ enough} from the previously observed ones (in either content or language usage).
That is, if the dataset of interest evolves in a non-stationary fashion over time.
The question is then how to characterize such non-stationary features.
%
%

We introduce and study four datasets that we extract from \textsc{Reddit} --- the news aggregator, content rating and discussion website.
The goal we seek is to predict the popularity of future post, given the history (and content) of past ones.
We are thus principally concerned with the two questions, namely (i) do these datasets exhibit strong enough distribution shift across time to affect the performance of LPLM, fine-tuned on the history of past posts? and, if so, (ii) how can we deal with such natural domain change problems?
Each of the \textsc{Reddit} datasets consists, as usual, of a single sequence of document collections (i.e. each time point in the sequence consists of an aggregate of documents with a given timestamp).
%
This practical aspect
entails, in particular, that the inference of representations capturing their relevant dynamic components (i.e. the non-stationary features from above) --- if at all present --- should be done via low capacity models.
%
%
Bayesian generative models for sequences, such as Kalman filters or Gaussian processes, are good examples well suited for such a task,
%
%
and Figure~\ref{fig:dynamics_of_wallstreetbets} illustrates this point.
In the first row of the figure we report the 30 most frequent words from all documents collected within the first, middle and last 8-week windows of one of our datasets, which spans about one year in total.
There is no discernible change between these three time points --- in this representation of the data ---
and one might jump to the conclusion that the dataset shows no dynamics (or that the data distribution is stationary). 
In the second row, however, we report how the proportions of three randomly chosen topics, inferred via a neural variational variant of Dynamic Topic Models (DTM) \cite{blei2006dynamic}, changes as time evolves. 
Some of the dynamic features of the dataset are now evident in this representation.
Note that the last row in the figure shows the top 30 words associated with each of the topics in the second row.

Our first contribution is to (empirically) show that LPLM, the likes of BERT \cite{devlin2018bert} and ROBERTA \cite{liu2019roberta}, fine-tuned on histories of past \textsc{Reddit} posts, display average performance drops of about 88$\%$ (in the best case!) when predicting the popularity of future posts.
In sharp contrast, we shall observe that LPLM perform very well on test sets extracted from the history of past posts.
This result thus responds affirmatively to our first question above. 

Our second and main contribution consists of a simple methodology to deal with the kind of temporal distribution shift we observe in \textsc{Reddit}.
Indeed, we strive to retain the expressiveness of deep neural language models (NLM) for treating the low-level word statistics composing the posts, while deploying DTM for encoding the kind of high-level document sequence dynamics shown in Figure~\ref{fig:dynamics_of_wallstreetbets}.
If one interprets the inferred topics as representing the \textit{domains} of the dataset, their inferred dynamics can, at least in principle, account for (some aspects of) the temporal domain changes present in the dataset.
Note that taking the view of topics as domains to deal with distribution shift problems has also been taken in the past, albeit in static settings \cite{hu-etal-2014-polylingual, guo-etal-2009-domain, oren-etal-2019-distributionally}.
A bit more in detail, our approach consists of mainly three components.
First, we use neural variational DTM to infer the time- and document-dependent proportions of a set of latent topics that best describe the data collections.
We represent this set of topics via learnable topic embeddings.
Second, we deploy NLM to encode the word sequences composing the posts into sequences of contextualised word representations. 
Third, we modify a recently proposed attention mechanism \cite{pmlr-v108-wang20c} to construct temporal post representations sensitive to the temporal domain changes.
These depend on both the NLM word representations of the post in question and the history of the dataset, as represented by the latent topics and their time-dependent proportions.
The resulting representations can be used to predict the popularity of future posts.

Below we show our approach significantly outperforms LPLM. 
Indeed, our models display performance drops of only about 40$\%$ in the worst cases (2$\%$ in the best ones) when predicting the popularity of future posts, while using only about 7$\%$ of the total number of parameters of LPLM
and providing interpretable representations that offer insight into real-world events, like the GameStop short squeeze of 2021.

%% file: related_work.tex
As discussed above, our methodology and the problems we tackle with it merge concepts from (i) dynamic topic models, (ii) neural models that encode both natural language and time information for supervised downstream tasks, and (iii) studies of natural distribution shifts across time and their effect on language models.

\textbf{Dynamic topic models}. The seminal work of \citet{blei2006dynamic} introduced the Dynamic Topic Model (DTM), which uses a state space model on the natural parameters of the distribution representing the topics, thus allowing the latter to change with time.
The DTM methodology was first extended by \citet{10.5555/3020488.3020493} to a nonparametric setting, via the correlation of Dirichlet process mixture models in time.
Later \citet{wang2012continuous} replaced the discrete state space model of DTM with a Diffusion process, thereby extending the approach to a continuous time setting.
\citet{jahnichen2018scalable} further extended DTM by introducing Gaussian process priors that allowed for a non-Markovian representation of the dynamics.
Other recent work on dynamic topic models is that of \citet{DBLP:journals/corr/abs-1805-02203}.
Another line of research leverages neural networks to improve the performance of topic models, the so-called neural topic models \cite{miao2016neural, srivastava2017autoencoding, zhang2018whai, dieng2020topic} which deploy neural variational inference \cite{kingma2013auto} for training. 
Within this line, the neural model of \citet{dieng2019dynamic} represent topics as dynamic embeddings, and model words via categorical distributions whose parameters are given by the inner product between the static word embeddings and the dynamic topic embeddings. As such, this model corresponds to the dynamic extension of \citet{dieng2020topic}.


\textbf{Dynamic language models for supervised downstream tasks}.
Incorporating temporal information into (neural) models of text is key to capture the constant state of flux typical of streaming text datasets, the likes of news article collections or social media content.
Very early in the game, \citet{yogatama-etal-2014-dynamic} considered using temporal, non-linguistic data to condition $n$-gram language models and predict economics-related content at a given time.
More recently \citet{dynamic_language_model_} learned, via recurrent neural models, hidden variables encoding time information, which are then used both to condition neural language models and in classification tasks.
Similarly, \citet{9206768} leveraged recurrent neural point process models to infer dynamic representations that help model both content and arrival times of Yelp reviews.
Other recent work have also used both temporal and text information, but to predicting review ratings \cite{wu2016joint, 10.1007/978-3-658-36295-9_10} instead.
Yet another very recent direction involves defining protocols to update the parameters of LPLM when applied to streaming data \cite{amba2021dynamic, liska2022streamingqa},
with the work of \citet{liska2022streamingqa}, in particular, also focusing on temporal distribution shifts.

Different from all these works, we use DTM to infer the dynamic components of our corpora, and attention mechanisms to connect them with neural language models for regression tasks.

\textbf{Natural distribution shifts across time in NLP}. 
WILDS, the benchmark introduced by \citet{pmlr-v139-koh21a}, is a very recent dataset collection which explores different types of real-world distribution shifts. 
Section 8 in \cite{pmlr-v139-koh21a} focus particularly on distribution shifts in NLP and we refer the reader to it for additional details and references. 
Additional to these are the aforementioned works which use topic models in static settings, to tackle domain change problems \cite{hu-etal-2014-polylingual, guo-etal-2009-domain, oren-etal-2019-distributionally}, as well as those works which report performance drops of LPLM under distribution shifts \cite{ma-etal-2019-domain, hendrycks-etal-2020-pretrained, zhou-etal-2021-contrastive, chawla2021quantifying}. 
Interestingly enough, WILDS includes in Appendix F a section about \textit{temporal} distributions shift for review data.
Nevertheless, they report only modest performance drops.
We do not find these results surprising, for review data typically deals with items (e.g. products, restaurants, etc) whose basic features change moderately with time.

To the best of our knowledge our work is the first to present a NLP dataset displaying temporal distribution shifts on which the perfomance of LPLM drops to a significant degree.



%% file: model.tex
\begin{figure*}[!t]
  \centering
  \includegraphics[keepaspectratio,width=\textwidth]{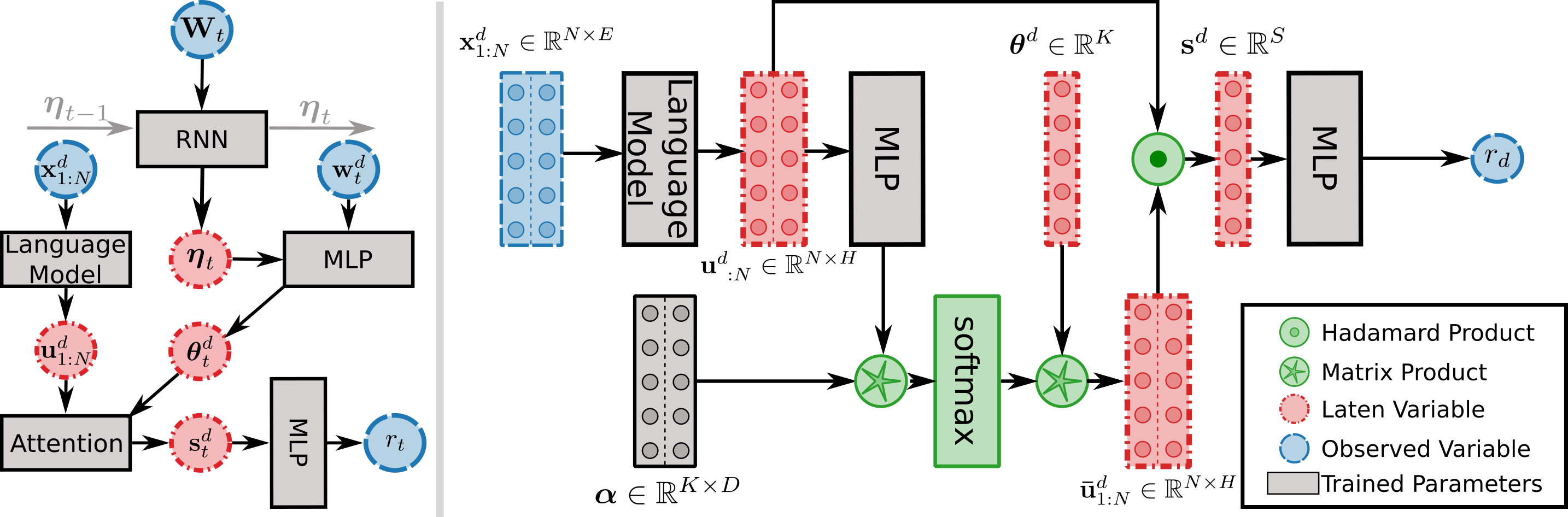}
    \caption{Left: \texttt{D-TAM-GRU} model. The input to the model is the document's word sequence and BoW representation, the output is a continuous label. The model consists of four components: (i) the DTM, (ii) the NLM encoder, (iii) the attention module and (iv) a regressor module, which takes the output from the attention module and predicts a label. Right: Detailed view of the \textit{Attention Module}. This module takes as input the document's word representations from the NLM and the topic proportion from the DTM.}
\label{fig:dynamic-model}
\end{figure*}
In this section we propose a simple methodology to deal with a class of distribution shift that naturally arises in corpora collected over long periods of time.
Suppose we are given an ordered collection of corpora $\mathcal{D}=\{D_1, D_2, \dots, D_T\}$,
so that the $t$th corpus $D_t$ is composed of $N_t$ documents (\textsc{Reddit} posts), all received within the $t$th time window. 
Let the $d$th document in $D_t$ be defined by the tuple $(\*w_{t,d}, \*X_{t,d}, r_{t,d})$, 
where $\*w_{t,d}$ denotes the Bag-of-Words (BoW) representation of the document, 
$\*X_{t,d} = (\*x^{t,d}_1, \dots,\*x^{t,d}_M)$ denotes the sequence of words comprising the document and 
$r_{t,d}$ labels the document's rating.
Similarly, let $\*W_t$ denote the BoW representation for the entire document set within $D_t$.
Given a new document $d'$ at time $T+1$ --- that is, given $\*X_{T+1, d'}$, $\*w_{T+1, d'}$ ---  the task is to predict the rating $r_{t+1, d'}$. 

Our approach takes the perspective that latent topics, understood as word aggregates grouped together by means of word co-occurrence information within the corpora, can be understood as representing the \textit{domains of the dataset}. 
To allow the distribution of topics within the documents to change with time allows, at least in principle, to model domain changes within the collection as time evolves. 
We shall use such temporal information to weight the importance of the words composing the new document $\*X_{T+1,d'}$, with respect to the topic (domain) proportions in that document, \textit{at time} $T+1$, by means of an attention mechanism.

The model thus requires the introduction of two components, namely (i) a neural variational DTM and (ii) an attention module that leverages the representations obtained by both the DTM and a low-level NLM, to create a temporal \textsc{Reddit} post representation.  
The resulting representation can then be used by a neural regressor model to predict $r_{t+1, d'}$.
%

In the following we introduce all the different components of the model in detail.

\subsection{Neural Variational Dynamic Topic Model}
\label{sec:DTM}
Let us suppose the corpora collection $\mathcal{D}$ is described by a set of $K$ unknown topics (domains).
We then assume there is a sequence of \textit{global} hidden variables $\+\eta_1, \dots, \+\eta_T \in \mathbb{R}^{\mbox{\footnotesize dim}(\eta)}$ which encodes how the topic proportions change among the corpora as time evolves (i.e.~as one moves from $D_t$ to $D_{t+1}$). 
We also assume there is a \textit{local} hidden variable $\+\zeta_{t, d} \in \mathbb{R}^{\mbox{\footnotesize dim}(\zeta)}$, conditioned on $\+\eta_t$, which encodes the content of the $d$th document in $D_t$, in terms of the $K$ topics.

\textbf{Generative model}. 
Let us denote with $\psi$ the set of parameters of our generative model.
%
%
We generate the $d$th document in $D_t$ by first sampling the topic proportions $\+\theta_{t,d} \in [0, 1]^K$ as follows
\begin{eqnarray}
\+\eta_t &\sim& \mathcal{N}\left(\boldsymbol{\mu}^{\eta}_\psi(\+\eta_{t-1}), \delta \, \mathbf{I}\right), \label{eq:transition_eta}\\
\+\zeta_{t, d} & \sim & \mathcal{N}\left(\*W^{\zeta}_{\psi} \, \+\eta_t + \*c^{\zeta}_{\psi}, \, 1\right), \\
\boldsymbol{\theta}_{t, d} & = & \text{softmax}(\*f_{\psi}^{\zeta}(\+\zeta_{t, d})), \label{eq:gen-model-base}
\end{eqnarray}
where $\*f_{\psi}^{\zeta}: \mathbb{R}^{\mbox{\footnotesize dim}(\zeta)} \rightarrow \mathbb{R}^K$ is neural network with parameters in $\psi$ 
and $\*W_{\zeta} \in \mathbb{R}^{\mbox{ \footnotesize dim}(\zeta) \times \mbox{ \footnotesize dim}(\eta)}, \*c_{\xi}\in\mathbb{R}^{\mbox{ \footnotesize dim}(\zeta)} \subset \psi$ are trainable parameters. 
Furthermore, and just as in Deep Kalman Filters \cite{krishnan2015deep}, $\+\eta_t$ is Markovian and evolves under a Gaussian noise with mean
$\boldsymbol{\mu}^{\eta}_{\psi}: \mathbb{R}^{\mbox{\footnotesize dim}(\eta)} \rightarrow \mathbb{R}^{\mbox{\footnotesize dim}(\eta)}$, defined via a neural network with parameters $\psi$, and variance $\delta$.  
The latter being a hyperparameter of the model.
Finally, we choose the prior $\+\eta_1 \sim \mathcal{N}(0, 1)$.
%

Once we have $\+\theta_{t,d}$ we generate the corpora sequence by sampling 
\begin{eqnarray}
z_{t, d, n} &\sim& \mbox{Categorical}(\boldsymbol{\theta}_{t, d}), \\
w_{t, d, n} &\sim& \mbox{Categorical}(\boldsymbol{\beta}_{z_{t, d, n}}),
\end{eqnarray}
where $z_{t, d, n}$ is the time-dependent topic assignment for $w_{t, d, n}$, which labels the $n$th word in document $d \in D_t$, and $\boldsymbol{\beta}\in\mathbb{R}^{K \times V}$ is a learnable topic distribution over words. We define the latter as
\begin{equation}
    \boldsymbol{\beta} = \mbox{softmax} (\+\alpha \otimes \+\rho),
    \label{eq:topic_distribution}
\end{equation}
with $\+\alpha \in \mathbb{R}^{K \times E}, \+\rho \in \mathbb{R}^{V \times E}$ learnable topic and word embeddings, respectively, for some embedding dimension $E$, and $\otimes$ denoting tensor product.

\textbf{Inference model}.  We approximate the true posterior distribution of the hidden variables with a variational (and structured) posterior of the form
\begin{equation}
q_{\varphi}(\+\eta_{t}, \+\zeta_{t, d} | \*w_{t, d}, \*W_{1:T}) = \prod_{t}^T q_{\varphi}(\+\eta_{t} | \+\eta_{1:t-1}, \*W_{1:T}) \times \prod_{d}^{N_t} q_{\varphi} (\+\zeta_{t, d} | \*w_{t, d}, \+\eta_t) 
\label{eq:posterior_distribution}
\end{equation}
where  $\*W_{1:T} = (\*W_1, \dots, \*W_T)$ is the ordered sequence of BoW representations for the corpus collection and $\varphi$ labels the variational parameters.
The posterior distribution over the local variables is chosen to be Gaussian
\begin{equation}
q_{\varphi}(\+\zeta_{t, d}|\*w_{t, d}, \+\eta_t)  =  \mathcal{N}(\+\mu^{\zeta}_{\varphi}[\*w_{t, d}, \+\eta_t], \+\sigma^{\zeta}_\varphi[\*w_{t, d}, \+\eta_t]),
\label{eq:local_posterior}
\end{equation}
with $\+\mu^{\zeta}_{\varphi}$ and $\+\sigma^{\zeta}_{\varphi}$ neural networks with parameter $\varphi$. 
Likewise, the posterior distribution over the global variables is also Gaussian, but now depends not only on the latent variables at time $t-1$, but also on the entire sequence of BoW representations $\*W_{1:T}$. 
Explicitly we write
\begin{equation}
q_{\varphi}(\+\eta_t|\+\eta_{t-1},\*W_{1:T}) =  \mathcal{N}(\+\mu^{\eta}_{\varphi}[\+\eta_{t-1}, \*h^{\eta}_T],\+\sigma^{\eta}_{\varphi}[\+\eta_{t-1}, \*h^{\eta}_T]),
\label{eq:posterior_global_1}
\end{equation}
where $\+\mu^{\eta}_{\varphi}, \+\sigma^{\eta}_{\varphi}$ are too given by neural networks and $\*h^{\eta}_t$ is a recurrent representation encoding the sequence $\*W_{1:T}$. Indeed,
\begin{equation}
\*h^{\eta}_t = g^{\eta}_{\varphi}(\*W_t, \*h^{\eta}_{t-1}),
\end{equation}
with $g_{\varphi}$ a neural model for sequence processing (like e.g. a GRU \cite{cho2014properties}). 
Figure \ref{fig:dynamic-model} illustrates the architecture of the complete model.
%

\textbf{Training objective}.
To optimize the DTM parameters $\{\psi, \varphi\}$ we minimize the variational lower bound on the logarithm of the marginal likelihood $p_{\psi}(w_{t, d, n}| \+\beta)$. Following standard methods \cite{bishop2006pattern}, the latter can readily be shown to be 
%

\begin{multline}
 \mathcal{L}[\boldsymbol{\beta}, \psi, \varphi]  = \sum_{t=1}^T \sum_{d=1}^{N_t}\Bigg(\mathbb{E}_{\left\{\+\zeta_{t, d}, \+\eta_{t}\right\}} \Big\{\log  p_{\psi}(\*w_{t, d}|\boldsymbol{\beta}, \+\zeta_{t,d},\+\eta_t) - \mbox{KL}\left[q_{\varphi}(\+\zeta_{t, d} | \*w_{t, d}, \+\eta_t); p_{\psi}(\+\zeta_{t, d} | \+\eta_{t}) \right] \Big\} \Bigg) \\ - \mbox{KL}\left[q_{\varphi}(\+\eta_1 | \*W_{1:T}); p(\+\eta_1)\right] -\sum_{t=2}^T \mbox{KL}\left[q_{\varphi}(\+\eta_t| \+\eta_{1:t-1}, \*W_{1:T}) ; p_{\psi}(\+\eta_t| \+\eta_{t-1}) \right],
 \label{eq:topic_model-loss}
\end{multline}
where KL labels the Kullback-Leibler divergence and $\+\beta$ is given in \eqref{eq:topic_distribution}.

\subsection{Neural Topic Attention Model}
%
Given a document $d$, composed of the word sequence $\*x_1, \dots, \*x_M$, the task is to predict its rating $r_{d}$.
One straightforward approach to this problem is to consider a (deep) neural language model to infer contextualized  word representations of the form
\begin{equation}
    \*u_1, \*u_2, \dots, \*u_M = s_{\rho}(\*x_1, \*x_2, \dots, \*x_M),
\end{equation}
where $s_{\rho}$ is any neural sequence processing model (as e.g. a GRU \cite{cho2014properties} or BERT model) with parameters $\rho$.

With the representations $\*u_1, \*u_2, \dots, \*u_M$  at hand one can define a summary representation $\*s_{d}$ (by e.g. averaging over the $\*u$'s or using the \textsc{CLS} token of BERT-like models) and use it as input to a neural regressor $\*f^r_{\rho}$ (also with parameters $\rho$) to predict $r_{d}$.
Yet, if the document $d$ is received at time $T+1$, i.e. out of the observation window on which the parameter $\rho$ was optimised, 
and if the dataset in question displays temporal domain changes, we might expect our simple model to underperform. 

Here we use the DTM of section~\ref{sec:DTM}, which is assumed to model the domain changes in the corpora, to define a temporal summary representation sensible to the natural distribution shifts of the dataset.
Indeed, we follow \citet{pmlr-v108-wang20c} and use  an attention mechanism to construct $\*s_{d}$ as follows
\begin{equation}
    \*s_{t, d} = \sum_j^M\sum_i^K (\theta_{t,d}^i - \delta) \, \text{softmax}(\text{MLP}(\*u_j)^T \cdot \+\alpha_i) \, \odot\*u_j,
    \label{eq:attention_topic}
\end{equation}
where the softmax function is taken with respect to the word sequence, 
the $\+\alpha_i$ label the set of $K$ global topic embeddings and the $\theta_{t,d}$ denote the time- and document-dependent topic proportions, as defined in Eq.~\ref{eq:gen-model-base}.

It follows that the proposed document representation is nothing but the sum of the projections of each of the document's word representations onto topic space, weighted by the \textit{time-dependent} topic proportions of each dimension.
It can also be understood as an attentive representation in which each word is queried by the weighted topics.

In what follows we let $\*s_{\rho}$ be modeled by a GRU network \cite{cho2014properties} and name the model thus defined \texttt{D-TAM-GRU}.
Our proposed framework is shown in Figure \ref{fig:dynamic-model}. 



\subsection{Training Objective}

To optimize the complete set of model parameters $\{\psi, \varphi, \+\rho \}$ we minimize the objective
\begin{equation}
    \mathcal{L}=\mathcal{L}_{TM}[\boldsymbol{\beta}, \psi, \varphi] + \alpha_y\mathcal{L}_{r}[\boldsymbol{\rho}],
    \label{eq:full-loss}
\end{equation}
where $\mathcal{L}_{TM}$ denotes the DTM loss (defined in Eq.~\ref{eq:topic_model-loss}) and  $\mathcal{L}_{r}$ denotes the regression loss. 
The latter is defined as the root-mean-squared error between the target rating $r_d$ and the prediction of a regressor model, that is
\begin{equation}
    \mathcal{L}_{r}[\boldsymbol{\rho}] = \text{RMSE}(r_d, \*f^{r}_{\rho}\left(\*s_d\right)),
    \label{eq:label_prediction-loss}
\end{equation}
with $\*f^{r}: \mathbb{R}^{\mbox{\footnotesize dim}(s)} \rightarrow \mathbb{R}$ a neural network and $\*s_d$ defined by Eq.~\ref{eq:attention_topic}.

\subsection{Prediction}

In order to predict the rating $r_{T+N, d}$ of a new document, $N$ steps into the future, given its word sequence $\*X_{T+N, d}$, we rely on the generative process of our model albeit conditioned on the past. 
Essentially one must generate Monte Carlo samples from the posterior distribution and propagate the global latent representations into the future with the help of the prior transition function Eq,.~\ref{eq:transition_eta}.
This procedure is depicted on the conditional predictive distribution (for a single step) of our model
\begin{multline}
p(r_{T+1,d}|\mathcal{D},\*X_{T+1, d}) = \int p(r_{T+1,d}| \+\eta_{T+1},\+\zeta_{T+1, d}, \*X_{T+1, d}) \\ \times p(\+\zeta_{T+1, d}|\+\eta_{T+1}) p(\+\eta_{T+1}|\+\eta_{T}) p(\+\eta_{1:T}|\mathcal{D})d\+\eta_{1:T+1} d\+\zeta_{T+1, d},
\end{multline}
where $p(\+\eta_{1:t}|\mathcal{D})$ is the exact posterior over the dynamical global variables, which we approximate with our variational expression Eq.~\ref{eq:posterior_global_1}, and where 
\begin{equation}
    p(r_{T+1,d}| \+\eta_{T+1},\+\zeta_{T+1, d}, \*X_{T+1, d}) = \delta(r_{T+1,d} - \*f_{\rho}^r(\*s_{T+1, d}[\+\eta_{T+1},\+\zeta_{T+1, d}, \*X_{T+1, d}])),
\end{equation}
with $\*f_{\rho}^r$ the neural regressor,  $\*s_{T+1, d}$ defined in Eq.~\eqref{eq:attention_topic} and $\delta$ the Dirac delta function.
\input{tables/test/all_results}

%% file: tables/test/all_results.tex
\begin{table}[h!]
\centering
\small
\caption{In-distribution results (Results in Test set defined in  \textsc{up-to-date} data), which violate causality, we use information from the future to predict the past. For each of the number of topics column we underline the best models within and with boldface the best models overall.}
\label{tab:test}
\begin{tabular}{ll>{\columncolor[HTML]{f1f0f0}}r>{\columncolor[HTML]{f1f0f0}}r>{\columncolor[HTML]{f1f0f0}}rrrr>{\columncolor[HTML]{f1f0f0}}r>{\columncolor[HTML]{f1f0f0}}r>{\columncolor[HTML]{f1f0f0}}r}
\toprule
 & \textbf{Model} & \multicolumn{3}{c}{\cellcolor[HTML]{f1f0f0}\textbf{R2$\uparrow$}} & \multicolumn{3}{c}{\textbf{PPL-DC$\downarrow$}} \\
 & & \textbf{25} & \textbf{50} & \textbf{100} & \textbf{25} & \textbf{50} & \textbf{100} \\
\midrule
\multirow{8}{*}{\rotatebox[origin=c]{90}{\textsc{AskScience}}} & \texttt{MLP} & 0.0007 & & & & &  \\
                                                                & \texttt{BERT} &  \underline{0.0712} & & & & &  \\
                                                                & \texttt{RoBERTa} & 0.0484 & & & & &  \\
                                                                & \texttt{TAM-GRU} & 0.0053 & 0.0555 & 0.0449 & 1615 & \underline{1616} & 1561 \\
                                                                & \texttt{TAM-BERT} & 0.0103 &  \textbf{\underline{0.0967}} &  \underline{0.0867} & \textbf{\underline{1462}} & 1647 & \underline{1549} \\
                                                                & \texttt{TAM-RoBERTa} & -0.0175 & -0.0169 & 0.0295 & 3290 & 1770 & 1612 \\
                                                                \cmidrule(lr){2-11} 
                                                                & \texttt{D-ST} & -0.0175 & -0.0138 & -0.0175 &  2147 & 1745 & 1691  \\
                                                                & \texttt{D-TAM-GRU} & 0.0387 & 0.0386 & 0.0397 & 1795 &  1813 &  1723 \\

\midrule
\multirow{8}{*}{\rotatebox[origin=c]{90}{\textsc{Politics}}} & \texttt{MLP} & 0.6278 & & & & &  \\
                                                            & \texttt{BERT} & 0.6820 & & & & &  \\
                                                            & \texttt{RoBERTa} &  \textbf{\underline{0.7306}} & & & & &  \\
                                                            & \texttt{TAM-GRU} & 0.7063 & \underline{0.6918} & \underline{0.6847} & \textbf{\underline{1857}} & 2007 & 2022  \\
                                                            & \texttt{TAM-BERT} & 0.7268 & 0.6639 & 0.6802 & 2075 & \underline{1986} & 1997  \\
                                                            & \texttt{TAM-RoBERTa} & -0.0191 & 0.6329 & 0.6305 & 3548 & 1998 & \underline{1976} \\
                                                            \cmidrule(lr){2-11} 
                                                            & \texttt{D-ST} & 0.3370 & 0.3631 & 0.2372 &  2157 &  2147 & 2014  \\
                                                            & \texttt{D-TAM-GRU} & 0.6705 & 0.6892 & 0.6836 & 2149 & 2118 &  2051  \\
\midrule
\multirow{8}{*}{\rotatebox[origin=c]{90}{\textsc{The Donald}}} & \texttt{MLP} & 0.4162 & & & & &  \\
                                                                & \texttt{BERT} & \textbf{\underline{0.6674}} & & & & &  \\
                                                                & \texttt{RoBERTa} & 0.5290 & & & & & \\
                                                                & \texttt{TAM-GRU} & 0.3493 & 0.3772 & 0.3195 & \underline{1887} & \textbf{\underline{1873}} & \underline{1893}  \\
                                                                & \texttt{TAM-BERT} & 0.5965 & \underline{0.6537} & \underline{0.6659} & 2138 & 1895 & 2009  \\
                                                                & \texttt{TAM-RoBERTa} & 0.5992 & 0.4878 & 0.5906 & 2195 & 2066 & 2073 \\
                                                                \cmidrule(lr){2-11} 
                                                                & \texttt{D-ST} & -0.0014 & -0.0015 & -0.0012 & 2162 & 2121 & 1986  \\
                                                                & \texttt{D-TAM-GRU} & 0.4374 & 0.4342 & 0.4302 &  2186 &  2125 &  2143 \\
\midrule
\multirow{8}{*}{\rotatebox[origin=c]{90}{\textsc{Wallstreetbets}}} & \texttt{MLP} & 0.5862 & & & & &  \\
                                                                & \texttt{BERT} &  \underline{0.6850} & & & & & \\
                                                                & \texttt{RoBERTa} & 0.5484 & & & & &  \\
                                                                & \texttt{TAM-GRU} & 0.4961 & 0.5134 & 0.5037 & 1373 & 1436 & 1390  \\
                                                                & \texttt{TAM-BERT} & 0.5253 & 0.4057 &  \textbf{\underline{0.8293}} & \underline{1341} & \underline{1360} & \textbf{\underline{1325}}  \\
                                                                & \texttt{TAM-RoBERTa} & -0.0084 & \underline{0.5994} & 0.0218 & 1539 & 1474 & 1355 \\
                                                                \cmidrule(lr){2-11} 
                                                                & \texttt{D-ST} & -0.0038 & -0.0022 & 0.1823 & 1693 &  1497 &  1566  \\
                                                                & \texttt{D-TAM-GRU} & 0.4989 & 0.5340 & 0.5066 &  1993 & 1466 & 1541  \\
\bottomrule
\end{tabular}
\end{table}

%% file: dataset_experimental_setup.tex
\input{tables/dataset_stats}
In this sections we present detailed information about our proposed \textsc{Reddit} dataset for temporal distribution shifts analysis. 
Next, we describe the experimental setup that we use to train and evaluate the models and
introduce the baselines we compare against.
\begin{figure}[h!]
\centering
\includegraphics[width=0.7\textwidth]{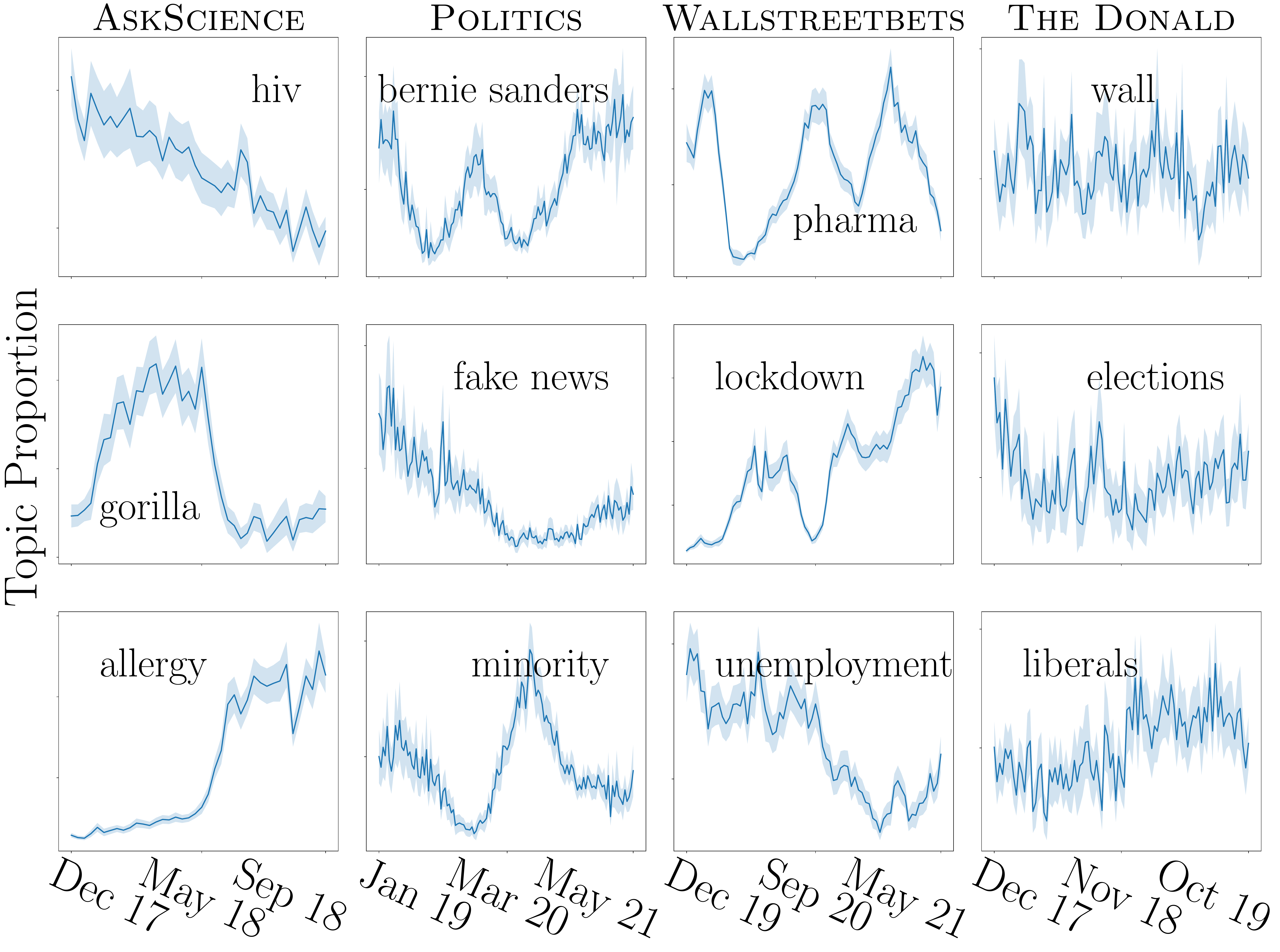} 
\caption{Time evolution of topics' proportion. The time series are obtained by taking the mean and two standard deviations of the activity $\theta_k$ of topic $k$ in all the documents for a given time step $t$. We present three randomly picked topics for each dataset. One could immediately notice that there is almost no dynamics for the \textsc{The Donald} dataset.}      
\label{fig:topic_dynamics}
\end{figure}
\subsection{Dataset}
\label{subsec:dataset}
In this work we propose a new  \textsc{Reddit}\footnote{\href{https://www.reddit.com}{https://www.reddit.com}} dataset for temporal distribution shifts analysis. 
\textsc{Reddit} is a news aggregator, content rating and discussion website. Users can post content on the site, like images, text links and videos, which are rated and commented by other users. The posts which are called \textit{submissions} are organized by subject in groups or \textit{subreddits}. According to Semrush, as of March 2022 \textsc{Reddit} is ranked as the 9th\footnote{\href{https://www.semrush.com/website/reddit.com/}{https://www.semrush.com/website/reddit.com/}} most visited website worldwide. 
We crawled the posts for the \textsc{AskScience}, \textsc{Politics}, \textsc{The Donald} and the \textsc{Wallstreetbets} subreddits. The \textsc{AskScience} is a subreddit in which science questions are posted and answered.
%
\textsc{Politics} is a subreddit where news and politics in the U.S. are discussed. 
\textsc{The Donald} was a subreddit where supporters of former U.S. president Donald Trump were initiating discussions. 
This subreddit was banned in June 2020 for violating \textsc{Reddit} rules. 
Lastly, the \textsc{Wallstreetbets} is a subreddit where stock trading is discussed. 
This subreddit played a major role in the GameStop short squeeze that caused losses\footnote{\href{https://www.bloomberg.com/news/articles/2021-01-25/gamestop-short-sellers-reload-bearish-bets-after-6-billion-loss}{https://www.bloomberg.com/news/articles/2021-01-25/gamestop-short-sellers-reload-bearish-bets-after-6-billion-loss}} for some U.S. firms in early 2021.

\subsection{Preprocessing} 
\label{subsec:preprocessing}
Crawled raw submission are preprocessed as follows. First, we remove all the submissions that are created by the automated system (e.g. when the author's name is \textit{AutoModerator}),
as well as submissions that have less than 20 words, and submissions containing text different than English. 
Next, we discretise the time interval into weekly time points (\textsc{AskScience} in monthly time steps) 
and randomly sub-sample 500 data points for each time point (1000 data points for \textsc{AskScience}).
We also use different preprocessing for the NLE and the DTM. 

\textbf{DTM Preprocessing}. We remove emojis, urls stop words, punctuation and spaces, and we lemmatize the words. 
We also remove all the words that appear in less than 5 documents, and we take the first 5000 words (in ascending order wrt. word frequency) as vocabulary, which we then use to create BoW vectors for each document.

\textbf{NLM Preprocessing}. Urls, spaces and other characters like HTML tags are removed and a maximum post length is defined (see Table \ref{tab:dataset_stats} for details).
Also all posts that have more than 50 000 comments are removed, and the target variable (i.e. the rating $r$ or number of comments) is scaled to have values between 0 and 1.

\textbf{Time Window and Out-of-Distribution Selection}.
After pre-processing, we first split each of the datasets into the \textsc{up-to-date} and \textsc{prediction} datasets. 
We create such a distinction to study temporal distribution shifts in the dataset.
Specifically, we take the last 20 time points as prediction (Out-of-distribution data) and the rest for the up-to-date posts (In-Distribution data). 
In this way we ensure that we do not train on documents that come from the future, which is what we actually want to model. That is, we do not violate causality. 
Likewise we ensure there is a clear distinction between past and future, which will allow us to uncover temporal distribution shifts, if present.
Table \ref{tab:dataset_stats} displays the specific timestamps which define the beginning and end of each dataset slice. 

Next, we split the \textsc{up-to-date} submissions randomly into train, validation and test sets (80\%, 10\%, 10\%, respectively).
Additionally, to evaluate the DTM on the document completion (i.e. generalization) task, we split the documents of the test set into two halves. 
The first half is used as input to the topic model; the second half is used to measure the document completion perplexity. 

Further statistics about the \textsc{Reddit} dataset (like e.g. histograms of the number of comments per post, etc) can be found in Fig.~\ref{fig:datasets_stats} in the Appendix.
\subsection{Baseline Models}
\label{subsec:baseline_models}
The baseline models are introduced in order of increasing complexity.
%
These models are generally composed of two modules, namely
(i) an \textit{encoder} module, which takes as input either the word sequence $\*X_{t,d}$ or the BoW $\*w_{t,d}$ of the document and outputs a summary representation $\*s_{d,t}$; (ii) a \textit{regressor} module, which takes as input the representation $\*s_{d,t}$ and predicts the rating $r_{t,d}$ of the document.

The simplest baseline model we consider defines both encoder and regressor as MLPs, and takes as input the BoW representation $\*w_{t,d}$ of the document.
We name it \texttt{MLP}. 
Next we introduce baselines with attention-based models \cite{vaswani2017attention} as encoder and MLPs as regressor. 
We use two attention-based encoder architectures:. \texttt{BERT} \cite{devlin2018bert} and \texttt{RoBERTa} \cite{liu2019roberta}.
The input to these models is the word sequence $\*X_{t,d}$ and we use their \textsc{CLS} embedding as input to the regressor module.
%
%
The third baseline is \texttt{TAM}, the neural topic attention model for supervised learning  proposed by \citet{pmlr-v108-wang20c}. 
\texttt{TAM} combines topic models and NLM to produce the representation $\*s_{d,t}$, just as in Eq.~\ref{eq:attention_topic} above, but with \textit{static} topic proportions. 
The original \texttt{TAM} version uses a GRU as NLM. We call this version \texttt{TAM-GRU}. 
We also extend this model by replacing the GRU with either \texttt{BERT} or \texttt{RoBERTa}. 
Accordingly we name these baselines \texttt{TAM-BERT} and \texttt{TAM-RoBERTa}, respectively. 
Finally, we use our DTM from section \ref{sec:DTM} as encoder module, and input the inferred local hidden variable $\+\zeta_{t,d}$ to the regressor, which here too is defined by an MLP.
We name this last baseline \texttt{D-ST}.
%

\subsection{Training and Evaluation Metrics}
\label{subsec:training}
We use grid search during training to find the best hyper-parameters of each model type. 
All models are trained on the training subset of the \textsc{up-to-date} submissions, and the validation subset is used for choosing the best hyper-parameters.
For all models that rely on a DTM module we use 25, 50 and 100 topics. Details regarding the values of other hyper-parameters can be found in the Appendix \ref{subsec:hyperparams}.

We quantify the performance of our models on the prediction tasks with the \textit{coefficient of determination} ($R^2$).
Additionaly, we also evaluate the performance of the topic models, whenever this are used, by means of
the \textit{predictive perplexity} (PPL-P),  the \textit{perplexity} on document completion (PPL-DC) \cite{rosen2012author} and the \textit{topic coherence} (TC) \cite{lau2014machine}.
%
%
Additional details about the metrics can be found in the appendix \ref{subsec:metrics}. We also provide the new proposed \textsc{Reddit} dataset and as well as the Torch \cite{NEURIPS2019_9015} implementation of the used models and scripts for downloading the data. 
%
%

%% file: tables/dataset_stats.tex
\begin{table*}[t!]
\tiny
\centering
\caption{Datasets' statistics.}
\label{tab:dataset_stats}
\begin{tabular}{
>{\columncolor[HTML]{EFEFEF}}c cccc}
\hline
                                    & \textsc{AskScience}          & \textsc{Politics}           & \textsc{The Donald}          & \textsc{Wallstreetbets}     \\ \hline
\textbf{maximum doc. size}                 & 120                 & 150                & 200                 & 500                \\
\textbf{\# of time steps train/prediction}   & 25/20 (months)                 & 124/20  (weeks)          & 93/25 (weeks)      & 72/25 (weeks)                   \\
\textbf{\# of train/test/prediction docs.} & 11 259/1 408/12 217 & 25 432/3 179/3 214 & 37 140/4 643/10 000 & 28 297/3 538/9 882 \\
\textbf{vocabulary size - TM/LM}    & 4 968/9 108         & 5 000/12 473       & 4 996/22 380        & 4 998/23 438       \\ 
\textbf{start-end train/prediction time}    & 01.01.18-01.01.20/30.09.21         & 01.01.19-11.05.21/30.09.21       & 01.01.2018-06.10.19/29.02.20       & 01.01.20-13.05.21/30.09.21        \\ \hline
\end{tabular}
\label{tab:dataset_stats}
\end{table*}

%% file: results_and_discussion.tex
\input{tables/predict/all_results}
In this section we discuss our results on the task of predicting the popularity of future submissions (posts) on the \textsc{Reddit} platform, by predicting the number of comments the submissions will receive.
As explained above, this task is defined by training all models on the \textsc{up-to-date} submission set and evaluating them on the \textsc{prediction} set, which consists of submissions received in the future.
%
%
\begin{figure}[h!]
\centering
\includegraphics[width=0.8\textwidth]{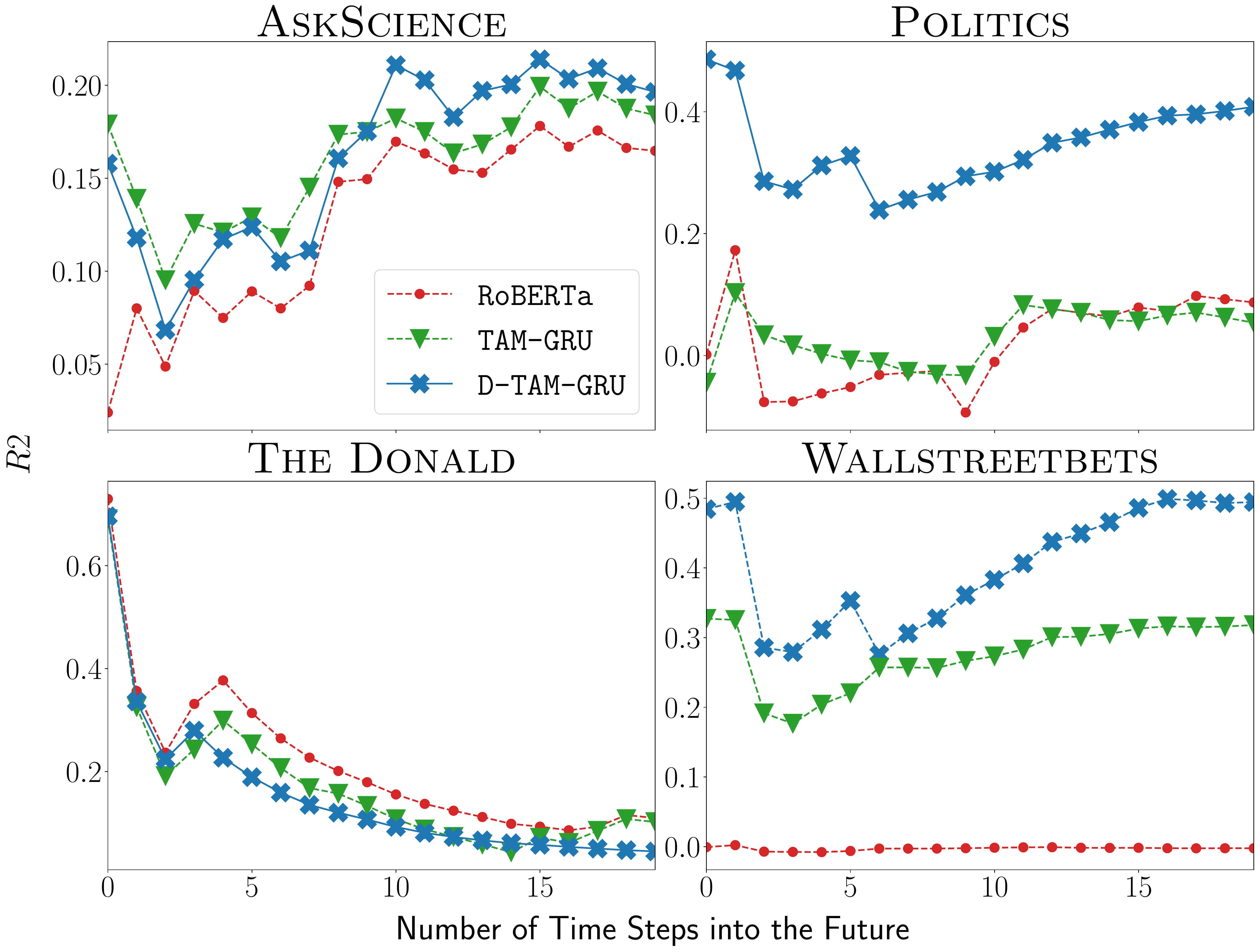} 
\caption{Cumulative average $R2$ score for each time step into the future obtained by using the predictions of the best transformer, the best \textit{static} TAM and the best \textit{dynamic} TAM models. Our models perform in all the datasets better than the static ones, except in the \textsc{The Donald} dataset which exhibits more stationary behavior, and in this case our model is not suitable. The rolling window size for calculating the cumulative average is 5 time steps.} 
\label{fig:prediction_timeseries}
\end{figure}
\begin{figure}[h!]
\centering
\includegraphics[width=0.8\textwidth]{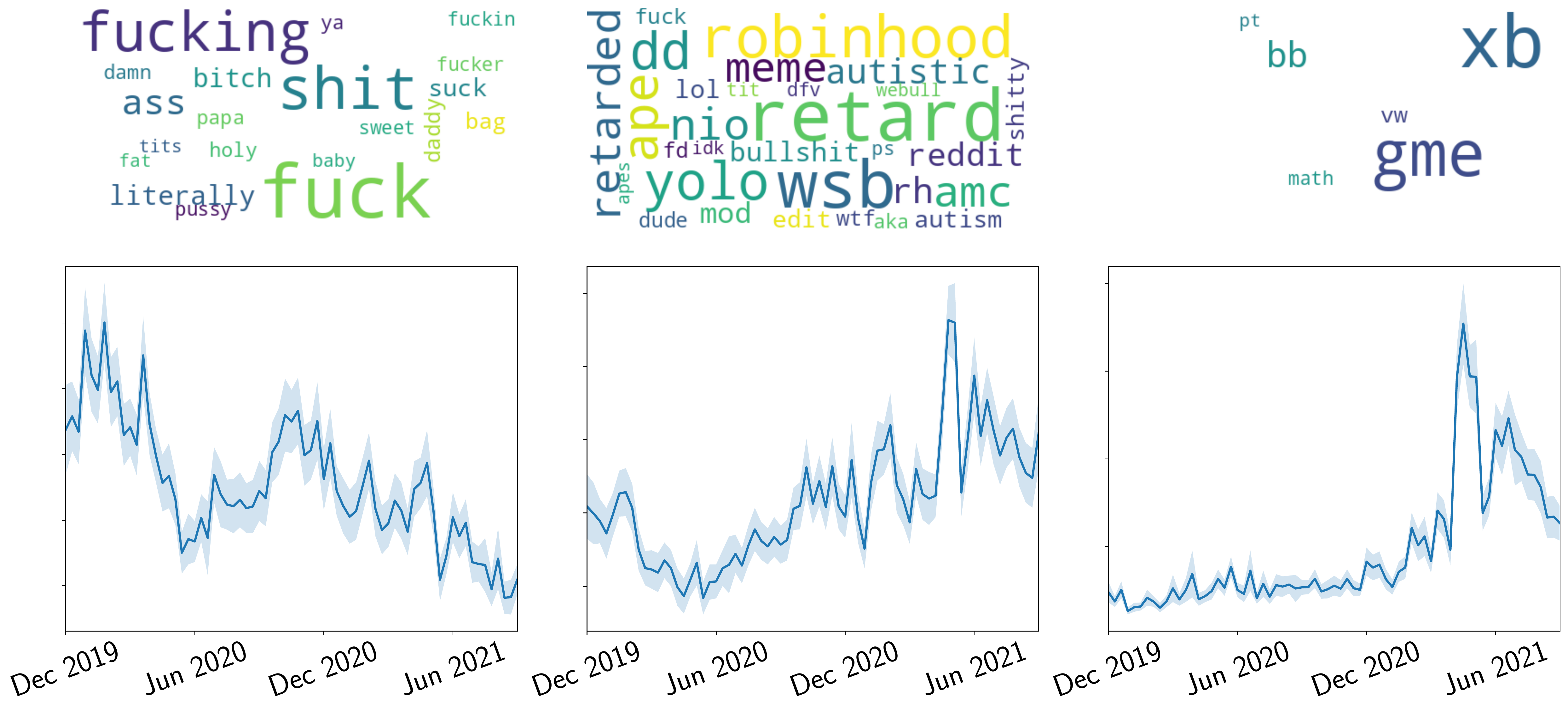}
\caption{Evolution of topic proportion in the \textsc{Wallstreetbets} dataset. Topics showing community culture (profane jargon) and the behavior of the population during to the "GameStop" short squeeze event.}
\label{fig:topic_evolution_wallstreetbets}
\end{figure}
One of the key takeaways of the present work is to highlight that LPLM, fine-tuned on the \textsc{up-to-date} submissions, fail at predicting the popularity of future posts.
Too see this let us first examine the results in Table~\ref{tab:test}, which show the performance of all models, evaluated on the test subset of the \textsc{up-to-date} submissions set.
That is, Table~\ref{tab:test} shows the In-distribution results.
Note how LPLM provide the best results in all subreddits, yielding $R^2$ scores which are 20\% to 30\% higher than the dynamic models, including our \texttt{D-TAM-GRU}.
In contrast, Table~\ref{tab:prediction} shows the Out-of-Distribution results, that is, the results of all models evaluated on the \textsc{prediction} set.
Comparing the performance of LPLM on the In-Distribution set against their performance on the Out-of-Distribution set, we observe performance drops of about 88$\%$ (in the best case!).
LPLM thus fail at predicting the popularity of future posts, and we understand these findings as being consequence of temporal distributions shifts between the \textsc{up-to-date} and \textsc{prediction} sets.

The second important observation we can make from Table~\ref{tab:prediction} is that, with the exception of those results in \textsc{The Donald} subreddit, \texttt{D-TAM-GRU} not only outperforms all LPLM, but also displays performance drops of only about 40$\%$ in the worst case (2$\%$ in the best ones) when predicting the popularity of future posts.
Thus, our simple methodology does in fact help dealing with the temporal distributions shifts of the \textsc{Reddit} dataset.
%
%
%
%
%
%
%

To understand why \texttt{D-TAM-GRU} performs as it does on \textsc{The Donald} subreddit, we studied the dynamic features inferred by our DTM on all subreddits.
Figure~\ref{fig:topic_dynamics} displays the time series for the topic proportions of three randomly selected topics from each dataset. 
Note how the topics exhibit a different range of dynamic behavior, accounting for seasonality, trendiness and bursty as well as simply random behavior. 
As a whole, the ability of the model to leverage such dynamic information in the prediction task strongly depends on the nature of these dynamical patterns, as well as the overall weight obtained by the topic attention mechanism. 
Indeed, one requires enough \textit{relevant} topics with dynamical information. 
In what can be thought of as a kind of distributed signal-to-noise ratio, we speculate that in order for
the prediction capabilities of NLMs to be improved by our approach, the dataset at hand must be such that there are enough topics with non-stationary behavior, i.e. topics that exhibit a distribution change over time, and that such topics are important for the prediction task, above other topics with stationary dynamical behavior (i.e. no change in time). 
\textsc{The Donald} dataset shows qualitative behavior that is overall stationary, as the topic proportions present mainly noisy behavior, which explains why \texttt{D-TAM-GRU} exhibits performance drops as large as those of the LPLM.

%
Finally, our ablation study also shows that (static) topic attention models expand the capabilities of LPLM. See Table~\ref{tab:prediction} and Figure~\ref{fig:prediction_timeseries}.
These findings comply with results in the literature, which indicate that topic models largely improve model performance under distribution shifts \cite{hu-etal-2014-polylingual, guo-etal-2009-domain, oren-etal-2019-distributionally}.
%

\textbf{Interpretability and \textsc{Wallstreetbets}}.
An added advantage of our model is the interpretable character of the representations inferred by our DTM, as we have seen in Figure~\ref{fig:topic_evolution_wallstreetbets}.
The \textsc{Wallstreetbets} subreddit has become a success story when it comes to the power the Web has to impact society,
as retail investors organized themselves in the platform to create major shifts in the stock market,
%
%
thereby playing a major role in short squeeze of the "GameStop" stock (an American video game retailer). 
This event can be directly observed in Fig.~\ref{fig:topic_evolution_wallstreetbets}-right, where the model inferred a rapid increase in the importance of a topic about ``gme" (ticker value of the ``GameStop" stock), previous to the sudden increase of the stock price by January 28, 2021. 
%
%
Now, due to the rapid increase of the stock price, some brokerages such as ``Robinhood" halted trading. 
The reaction of the community to this decision can also be observed in the rise of the topic shown in Fig.~ \ref{fig:topic_evolution_wallstreetbets}-middle, in which``Robinhood" is paired with derogatory jargon. 
New insights into the behavior of the population are uncovered by our model too. Fig.~\ref{fig:topic_evolution_wallstreetbets}-left shows how profane-related topic decays in importance with time, which means that the language used by the \textsc{Wallstreetbets} subreddit community is shifting. 

Beyond these qualitative results, the ability of our model to predict the popularity of posts, allows us to quantify the impact of several topics in the platform, as well as to predict popularity shifts within the user population. 
As \textsc{Wallstreetbets} continues to gain ground with retail investors, our methodology opens a window to quantitatively study possible future rises in the popularity of futures stocks in the \textsc{Reddit} platform.

\section{Conclusion}
In this work we studied the prediction capabilities of large pre-trained language models (LPLM) and showed that, for a newly introduced dataset with rich dynamic behavior, temporal distribution shifts produce a sharp drop in the performance of these LPLM.
We introduced a neural variational dynamic topic model with attention that outperforms LPLM and overcomes (some of) the difficulties created by the temporal distribution shifts.
Remarkably, our models use only about 7\% of the total number of parameters of LPLM and provide interpretable representations that offer insight into real-world events. 
%

%% file: tables/predict/all_results.tex
\begin{table*}[t!]
\small
\centering
\caption{\textsc{prediction}: Out-of-Distribution results which respect causality, the information of the past is used to predict the future. For each of the number of topics column we underline the best models within and with boldface the best models overall.}
\label{tab:prediction}
\begin{tabular}{ll>{\columncolor[HTML]{f1f0f0}}r>{\columncolor[HTML]{f1f0f0}}r>{\columncolor[HTML]{f1f0f0}}rrrr>{\columncolor[HTML]{f1f0f0}}r>{\columncolor[HTML]{f1f0f0}}r>{\columncolor[HTML]{f1f0f0}}r}
\toprule
 & \textbf{Model} & \multicolumn{3}{c}{\cellcolor[HTML]{f1f0f0}\textbf{R2$\uparrow$}} & \multicolumn{3}{c}{\textbf{PPL-P$\downarrow$}} & \multicolumn{3}{c}{\cellcolor[HTML]{f1f0f0}\textbf{TC$\uparrow$}} \\
 & & \textbf{25} & \textbf{50} & \textbf{100} & \textbf{25} & \textbf{50} & \textbf{100} & \textbf{25} & \textbf{50} & \textbf{100} \\
\midrule
\multirow{8}{*}{\rotatebox[origin=c]{90}{\textsc{Askscience}}} & \texttt{MLP} & 0.011$\pm$0.012 &  &  &  &  & &  &  & \\
                                                            & \texttt{BERT} & -1.029$\pm$0.756 &  &  &  &  & &  &  & \\
                                                            & \texttt{RoBERTa} & 0.105$\pm$0.158 &  &  &  &  & &  &  & \\
                                                            & \texttt{TAM-GRU} & 0.075$\pm$0.096 & 0.150$\pm$0.118 & 0.164$\pm$0.141 & 1525 & 1609 & 1502 & -0.32 & -0.38 & -0.35\\
                                                            & \texttt{TAM-BERT} & 0.054$\pm$0.103 & 0.134$\pm$0.171 & 0.154$\pm$0.170 & \underline{1370} & \underline{1257} & \textbf{\underline{1348}} & \underline{-0.06} & \textbf{\underline{-0.14}} & \underline{-0.23}\\ 
                                                             & \texttt{TAM-RoBERTa} & -0.019$\pm$0.008 & -0.019$\pm$0.008 & 0.092$\pm$0.099 & 2668 & 1867 & 1620 & -0.45 & -0.36 & -0.41 \\
                                                            \cmidrule(lr){2-11} 
                                                            & \texttt{D-ST} & 0.008$\pm$0.030 & -0.010$\pm$0.010 & -0.006$\pm$0.008 & 1914 & 1867 & 2031 & -0.34 & -0.45 & -0.59 \\
                                                            & \texttt{D-TAM-GRU} & \underline{0.152$\pm$0.105} & \textbf{\underline{0.196$\pm$0.175}} & \underline{0.173$\pm$0.161} & 1904 & 1844 & 1752 & -0.27 & -0.39 & -0.50\\
\midrule
\multirow{8}{*}{\rotatebox[origin=c]{90}{\textsc{Politics}}} & \texttt{MLP} & -2.507$\pm$8.977 &  &  &  &  &  &  &  &\\
                                                             & \texttt{BERT} & -4.569$\pm$14.48 &  &  &  &  &  &  &  &\\
                                                             & \texttt{RoBERTa} & 0.087$\pm$0.352 &  &  &  &  & &  &  & \\
                                                             & \texttt{TAM-GRU} & \underline{0.194$\pm$0.255} & -0.017$\pm$0.017 & -0.421$\pm$1.079 & \textbf{\underline{1486}} & \underline{1486} & 1877 & -0.52 & -0.63 & -0.27\\
                                                             & \texttt{TAM-BERT} & -0.031$\pm$0.077 & -0.091$\pm$0.047 & -0.362$\pm$0.885 & 2126 & 1992 & \underline{1759} & -0.58 & -0.56 & \textbf{\underline{0.07}}\\
                                                             & \texttt{TAM-RoBERTa} & -0.091$\pm$0.047 & 0.054$\pm$0.227 & 0.008$\pm$0.408 & 3368 & 2081 & 2109 & -0.36 & -0.63 & -0.56\\
                                                            \cmidrule(lr){2-11} 
                                                             & \texttt{D-ST} & -0.072$\pm$0.216 & -0.191$\pm$0.127 & -0.178$\pm$0.291 & 2394 & 2425 & 2248 & -0.26 & -0.50 & -0.36\\
                                                             & \texttt{D-TAM-GRU} & -0.016$\pm$0.015 & \textbf{\underline{0.408$\pm$0.221}} & \underline{0.129$\pm$0.169} & 2336 & 2250 & 2288 & \underline{0.01} & \underline{-0.21} & -0.31\\
\midrule
\multirow{8}{*}{\rotatebox[origin=c]{90}{\textsc{The Donald}}} & \texttt{MLP} & -0.009$\pm$0.010 &  &  &  &  &  &  &  &\\
                                                            & \texttt{BERT} & -0.047$\pm$0.029 &  &  &  &  &  &  &  &\\
                                                            & \texttt{RoBERTa} & \textbf{\underline{0.110$\pm$0.258}} &  &  &  &  &  &  &  & \\
                                                            & \texttt{TAM-GRU} & 0.052$\pm$0.226 & 0.084$\pm$0.250 & \underline{0.102$\pm$0.281} & 2199 & 2061 & \underline{1608} & -0.53 & -0.63 & 0.03\\
                                                            & \texttt{TAM-BERT} & 0.079$\pm$0.292 & \underline{0.094$\pm$0.286} & -0.116$\pm$0.072 & \underline{2048} & \textbf{\underline{1594}} & 2263 & -0.12 & \underline{-0.12} & \textbf{\underline{0.05}}\\
                                                            & \texttt{TAM-RoBERTa} & -0.116$\pm$0.072 & -0.036$\pm$0.024 & -0.032$\pm$0.203 & 6331 & 1870 & 2041 & -0.29 & -0.34 & -0.42 \\
                                                            \cmidrule(lr){2-11} 
                                                            & \texttt{D-ST} & -0.011$\pm$0.011 & -0.012$\pm$0.011 & -0.011$\pm$0.010 & 2144 & 2105 & 1858 & \underline{0.03} & -0.48 & -0.52\\
                                                            & \texttt{D-TAM-GRU} & 0.045$\pm$0.181 & 0.048$\pm$0.209 & 0.085$\pm$0.251 & 2131 & 2183 & 2064 & -0.04 & -0.34 & -0.29 \\
\midrule
\multirow{8}{*}{\rotatebox[origin=c]{90}{\textsc{Wallstreetbets}}} & \texttt{MLP} & 0.481$\pm$0.506 & &  &  &  & &  &  & \\
                                                               & \texttt{BERT} & -2.044$\pm$8.304 &  &  &  &  &  &  &  &\\
                                                               & \texttt{RoBERTa} & -0.002$\pm$0.009 &  &  &  &  &  &  &  &\\
                                                               & \texttt{TAM-GRU} & 0.318$\pm$0.253 & -0.017$\pm$0.015 & -1.913$\pm$7.801 & 1486 & 1657 & 1271 & -0.40 & -0.59 & -0.29\\
                                                               & \texttt{TAM-BERT} & -0.016$\pm$0.015 & 0.494$\pm$0.120 & 0.042$\pm$0.054 & \underline{1232} & \underline{1350} & \textbf{\underline{1171}} & \underline{0.17} & -0.32 & \textbf{\underline{0.14}} \\
                                                               & \texttt{TAM-RoBERTa} & -0.014$\pm$0.025 & -0.008$\pm$0.005 & -0.280$\pm$1.139 & 2384 & 1543 & 1518 & -0.39 & -0.57 & -0.51 \\
                                                               \cmidrule(lr){2-11} 
                                                               & \texttt{D-ST} & \underline{0.510$\pm$0.296} & -0.008$\pm$0.006 & 0.376$\pm$0.165 & 2491 & 1782 & 1576 & -0.07 & \underline{-0.30} & -0.44\\
                                                               & \texttt{D-TAM-GRU} & -0.016$\pm$0.015 & \underline{0.503$\pm$0.291} & \textbf{\underline{0.524$\pm$0.199}} & 2160 & 1700 & 1583 & 0.00 & -0.33 & -0.22\\
\bottomrule
\end{tabular}
\end{table*}

%% file: appendix.tex
\section{Experimental Setup}
\subsection{Model Training Setup}
\label{subsec:hyperparams}
We use grid search during training to find the best hyper-parameters for each model type. All models are trained on the training set and the validation set is used for picking the best hyper-parameters for each model type. Early stopping during training was used, where as metric for improvement is used root-mean-squared error (RMSE). For all the models the representation size that comes out from the encoder is 128, except for the transformer models where the size is 768. The encoder size for all the model is MLP with two hidden layers of size 256 and we use Dropout \cite{srivastava2014dropout} with probability of 0.3.
For the models that combine topic model and transformer as language model we use two optimizer, one for the topic model and regressor and the second one for the transformer. This is done because we use already pre-trained transformers models that only needed to be finetuned with small learning rate. SGD \cite{kiefer1952stochastic} was used as first optimizer and AdamW \cite{loshchilov2017decoupled} for training the transformers. In the next we will describe for all the models the parameter grid used in the grid search algorithm: (1) \texttt{MLP}: learning rate $[0.001, 0.0005]$; batch size $[32, 128]$; regressor size [linear layer, two hidden layers of size 256]; max number of epochs is $200$; (2) \texttt{BERT}, \texttt{ALBERT} and \texttt{RoBERTa}: learning rate $[1\mathrm{e}^{-5}, 1\mathrm{e}^{-6}]$; batch size $[8, 16]$; regressor size two hidden layers of size 256; max number of epochs is $20$; (3) \texttt{TAM-GRU}: learning rate $[0.001, 0.0005]$; batch size $[32, 64, 128]$; regressor size two hidden layers of size 256; max number of epochs is $200$; (4) \texttt{ATM-<TRANSFORMER>}: learning rate $[(1\mathrm{e}^{-3}, 1\mathrm{e}^{-6}), (5\mathrm{e}^{-4}, 1\mathrm{e}^{-6})]$; batch size $[16,32]$; (5) \texttt{D-ST} and \texttt{D-TAM-GRU}: learning rate $[0.001, 0.0005]$; batch size $[32, 128]$; for inference of the global variable $\+\eta_t$ (see \Eqref{eq:posterior_global_1}) we use LSTM \cite{lstm} with 4 layers with hidden size 400 and dropout 0.3; the transition function of the global variable (see \Eqref{eq:transition_eta}) is MLP with two hidden layers of size 64; the size of the $\+\eta_t$ is equal to the number of topics. All the models that have $\alpha_y$ we look in $[1, 100, 500, 1000]$ and values for $\delta$ we take $0.2, 0.1, 0.005$ for the 25, 50 and 100 topics respectively. Pre-traind GloVE \cite{pennington2014glove} word embeddings are used for the word representations $\+\rho$ in the topic model and as word embeddings in the case when the GRU is used as LM.
\subsection{Metrics}
\label{subsec:metrics}

In order to quantify the performance of our models, we first focus on two aspects, namely its prediction capabilities and its ability to generalize to unseen data. To test how well our models perform on a prediction task we compute the \textit{predictive perplexity} (PPL-P). To our knowledge this metric does not appear explicitly in the dynamic topic model literature, therefore we define this metrics as


%
%
\begin{equation}
\text{PPL-P} = \text{exp}\Big\{\mathbb{E}_{p_{\psi}(\+\Gamma_{T+1}|\+\Gamma_{T})}\mathbb{E}_{q_{\varphi}(\+\Gamma_{1:T}| \*W_{1:T})} 
\Big\{  \log  p_{\psi}(\*W_{T+1}|\+\Gamma_{T+1}) \Big\}\Big\},
\label{eq: predictive_preplexity}
\end{equation}
where $\+\Gamma_{t}$ labels the set $\{\+\eta_t, \+\zeta_{t,d} \}$.

In order to test the generalization capabilities we use three metrics namely, \textit{perplexity} (PPL-DC) on document completion, \textit{topic coherence} (TC) and \textit{coefficient of determination} ($R^2$). 
The document completion PPL is calculated on the second half of the documents in the test set, conditioned on their first half \cite{rosen2012author}. 
The TC is calculated by taking the average pointwise mutual information between two words drawn randomly from the same topic \cite{lau2014machine} and measures the interpretability of the topic.
\subsection{Dataset Statistics}
\begin{figure}[h]
\centering
 \begin{subfigure}[b]{0.9\columnwidth}
         \centering
         \includegraphics[width=\columnwidth]{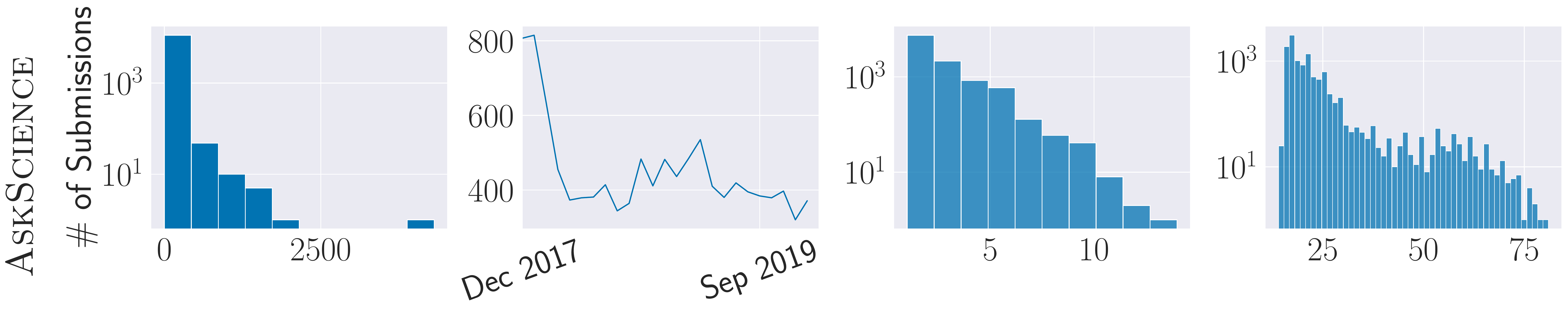}
         \label{fig:askscience_stats}
     \end{subfigure}
     \hfill
     \begin{subfigure}[b]{0.9\columnwidth}
         \centering
         \includegraphics[width=\columnwidth]{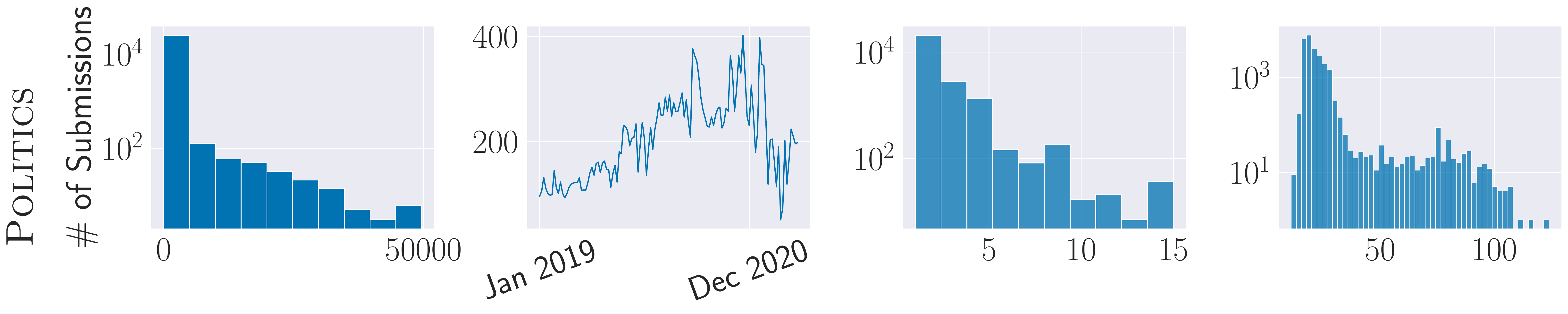}
         \label{fig:politics_stats}
     \end{subfigure}
     \hfill
     \begin{subfigure}[b]{0.9\columnwidth}
         \centering
         \includegraphics[width=\columnwidth]{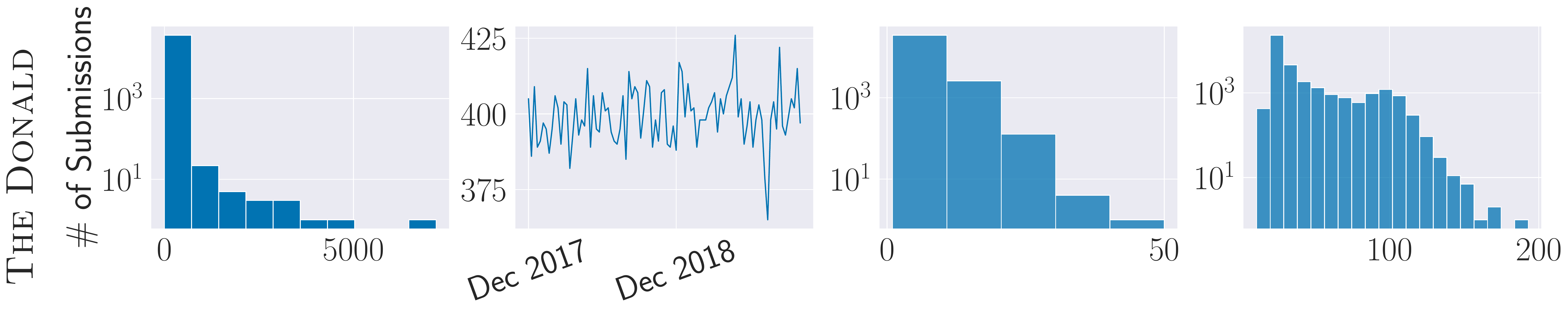}
         \label{fig:the_donald_stats}
     \end{subfigure}
     \hfill
     \begin{subfigure}[b]{0.9\columnwidth}
         \centering
         \includegraphics[width=\columnwidth]{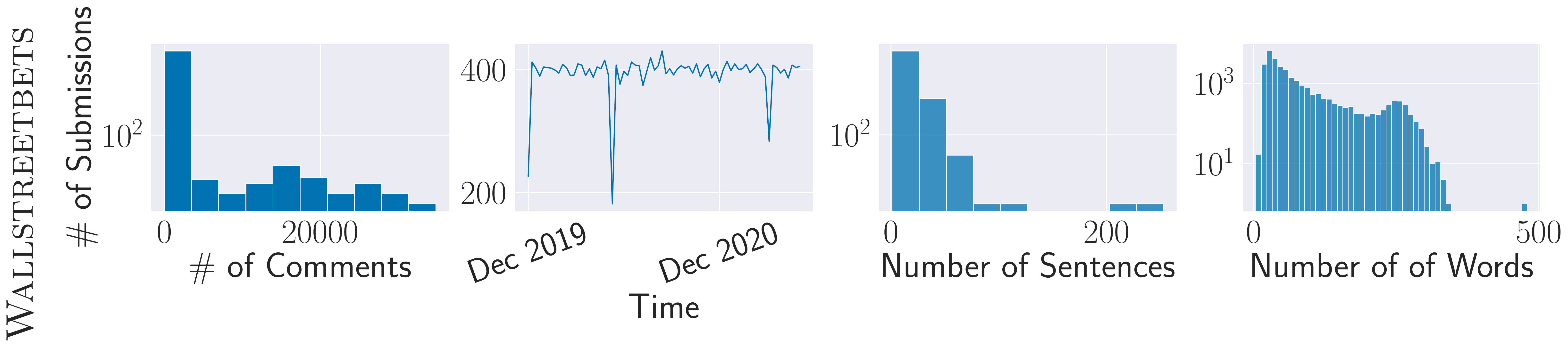}
         \label{fig:wallstreetbets_stats}
     \end{subfigure}
        \caption{\textsc{Reddit} datasets' statistics. In the first column are presented the histograms of the number of comments per submission. Next column, contains the number of submission per week (month in the case of \textsc{AskScience}). The last two columns contain the histograms of the number of sentences and number of words per submissions, respectively.}
        \label{fig:datasets_stats}
\end{figure}